# Sürü Robotları için Görüntü Tabanlı Mesafe, Açı ve Yönelim Sistemi Tasarımı

# Design of a Vision-Based Range, Bearing and Heading System for Robot Swarms

*Hamid Majidi Balanji*[1], *Emre Yılmaz*[1], *Ömer Çakmak*[1], *Ali Emre Turgut*[1]

[1]Makina Mühendisliği Bölümü
Orta Doğu Teknik Üniversitesi, Ankara
`{balanji.hamid, emre.yilmaz_02, omer.cakmak, aturgut}@metu.edu.tr`

## Özetçe

Sürü robotiğindeki esas problemlerden biri, sürü içerisindeki robotların birbirlerinin konumları hakkında nasıl bilgi sahibi olduğudur. Bu araştırmanın asıl amacı, sürü içerisindeki robotların çok amaçlı pasif işaretleyiciler kullanılarak birbirlerine göre olan mesafesi (range), açısı (bearing) ve yönelimi (heading) hakkında bilgi sahibi olmalarını sağlayan, basit ve uygun maliyetli, görüntü tabanlı bir algılama sistemi tasarlamaktır. Raspberry Pi ve Pi-Kamera ile donatılmış küçük bir Zumo robot, algoritmanın uyarlanması için kullanılmış ve robotların birbirlerine göre olan mesafesi, açısı ve yönelimi hakkında tek seferde doğru bilgi verebilecek simetrik olmayan desenlerle donatılmış pasif işaretleyiciler dizayn edilmiştir. Sistematik deneyler sonucu görüntülerden elde edilen veriler ile gerçek ölçümler karşılaştırılarak, gerçek değerlerin tespit edilebilmesi için gerekli ilişkiler elde edilmiş ve görüntü verilerini mesafe, açı ve yönelim değerlerine çeviren algoritmalar tasarlanmıştır. Algoritmaların güvenilebilirliği ve doğruluğu test edilmiş ve hata değerlerinin kabul edilebilir sınırlar içinde olduğu gösterilmiştir.

## Abstract

An essential problem of swarm robotics is how members of the swarm know positions of other robots. The main aim of this research is to develop a cost-effective and simple vision-based system to detect the range, bearing and heading of the robots inside a swarm using a multi-purpose passive landmark. A small zumo robot equipped with Raspberry Pi, PiCamera is utilized for the implementation of the algorithm, and different kinds of multipurpose passive landmarks with nonsymmetrical patterns, which give reliable information about the range, bearing and heading in a single unit, are designed. By comparing the recorded features obtained from image analysis of the landmark through systematical experimentation and the actual measurements, correlations are obtained and algorithms converting those features into range, bearing and heading are designed. The reliability and accuracy of algorithms are tested and errors are found within an acceptable range.

## 1. Giriş

Sürü robotik için, araştırmacılar tarafından şimdiye kadar birçok tanımlama yapılmıştır. Bunlardan en yaygını, sürü robotiğin, ortak bir görevi yerine getirmek için, çok sayıda nispeten basit robotun koordinasyonu olduğunu belirtmektedir [1]. Başka bir deyişle, çoklu robot sistemleri, ortak bir görevi tamamlamak üzere iş birliği yaparak çalışabilen sosyal böceklerin etkileşim davranışlarından ilham almaktadır. Ortaya atılan tanımlamadan bağımsız olarak, sürü robot sistemlerinin kendilerini diğer robot sistemlerinden ayıran dayanıklılık, esneklik ve ölçeklenebilirlik gibi ortak birtakım özellikleri paylaşmaları gerekmektedir.

Sürü içerisindeki robotların mesafe, açı ve yönelim değerlerinin gerçek zamanlı ölçümü sürü robotiğindeki en zorlu işlerden biri olarak görülebilir. Mesafe, açı ve yönelim değerlerinin bilinmesi robotların, zincir oluşumu, kendiliğinden montaj, eşgüdümlü hareket, çukurdan kaçma, yiyecek arama ve çarpışma önleme gibi görevlerde yer alabilmesine olanak tanır. Bu verilerin ölçümünde ultrasonik, kızılötesi, lazer, radar ya da görüntü algılama sistemleri gibi çeşitli araçlar kullanılabilmektedir. Mesafe ve açı algılama sistemleri genellikle bir araya getirme, sürü halinde hareket etme ve birlikte taşıma gibi sürü robot uygulamalarında kullanılmaktadır.

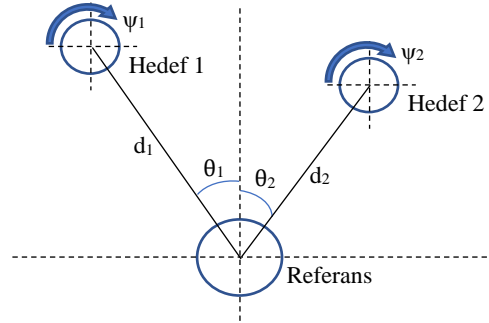

*Şekil 1:* Robotların birbirlerine göre olan mesafesinin, açısının ve yöneliminin gösterimi.

Robotlar arasındaki mesafe, her bir robottan hedefe ya da diğer robotlara olan uzaklık olarak tanımlanabilir. Robotlar arasındaki açı, robotların merkezlerini birleştiren doğrunun dikey eksenle yaptığı açıyı, robotun yönelimi ise her bir robotun bulunduğu konumdaki oryantasyonunu ifade etmektedir. Şekil 1 bu terimleri şematik olarak göstermektedir. Şekil 1'e göre, d, θ ve h, sırasıyla robotların birbirlerine göre olan mesafesini, açısını ve yönelimini ifade etmektedir. Bahsedilen mesafe, açı ve yönelim değerlerini doğrudan doğruya ölçebilen herhangi bir ticari araç olmadığından dolayı, araştırmacılar tarafından şimdiye kadar, bu terimlerin ölçümü konusunda çeşitli metotlar ortaya atılmıştır.

Kontrol teorisinde, mesafe, açı ve yönelim bilgileri sürü halinde hareket etme algoritmalarında kullanılmaktadır [2]. Tanner ve diğerleri [3][4] mesafe, açı ve yönelim bilgilerini sabit ve dinamik topoloji durumlarına bağlı olan açık uzay tabanındaki robotların sürü halinde hareket etmesini sağlayan, stabil bir kontrol kuralı tasarlamak için kullanmıştır. Cezayirli ve diğerleri [5] tarafından ortaya atılan bir sürü halinde hareket etme algoritmasında da komşu robotların mesafe ve açı bilgileri kullanılmaktadır. Yine benzer şekilde, Hanada ve diğerleri [6] de engellerle dolu bir alanda sürü halinde hareket etmek için bir algoritma ortaya atmıştır. Tasarlanan bu algoritmada deney alanındaki robotların mesafesi, açısı ve engellerin konumu tüm robotlar için bilinmektedir. Moshtagh ve diğerleri [7] sürü içerisindeki robotların yönelimini ayarlamak için görüntü tabanlı koordinasyon ve sürü halinde hareket etme algoritması geliştirmiştir. Bu algoritmada robotların açısının ölçümünde optik bir araç kullanılmıştır fakat yönelim değerleri dışarıdan herhangi bir araç ile ölçülmemiştir. Gervasi ve diğerleri [8] de benzer şekilde, sürü içerisindeki robotların mesafe ve açı değerlerine bağlı, bir sürü halinde hareket etme algoritması tasarlamıştır. Burada ise mesafe ve açı etkenleri bilinmektedir. Gregoire ve diğerleri [9] ise robotların sürü içerisinde tutarlı bir şekilde hareket etmelerini sağlamak için mesafe ve açı bilgilerini kullanmıştır. Kelly ve Keating [10] 10 tane robotun olduğu bir gruptaki robotların mesafe ve açı bilgilerini, özel yapım aktif bir kızılötesi sistem kullanarak hesaplamıştır. Spears ve diğerleri [11] tarafından 2 kızılötesi sensör modülü, robotların mesafesini ve açısını ölçmek için kullanılmıştır. Ortaya atılan algoritma, örgü oluşturma ve sürü halinde hareket etme uygulamalarına yöneliktir. Vaughan ve diğerleri [12] robotik bir çoban köpeği kullanarak ördekleri sürü halinde hareket ettirmek için bir metot geliştirmiştir. Kullandıkları metot mesafe ve açı bilgilerine gereksinim duymaktadır. Mesafe ve açı değerleri, ördekleri takip edip bilgileri lider çoban köpeğine gönderen bir tepe kamerası ile elde edilmektedir. Turgut ve diğerleri [2] Kobot sürü robot çalışmaları için, yakın mesafedeki robotların mesafe ve açı değerlerini ölçmek için kısa mesafe sensör sistemleri ile donatılmış bir platform tasarlamıştır.

James ve diğerleri [13] küçük boyutlu çoklu robotik sistemler arasındaki uzamsal koordinasyonu, 2.5 boyutlu kızılötesi mesafe ve açı sensörleriyle sağlamıştır. Bu sistem, sürü içerisinde pek çok robot olduğunda sorun yaratmıştır. Robotlar arasındaki mesafenin ve açının tespiti, hibrit kızılötesi ve yarım açı yoğunluğu ±25º olan 16TSAL4400 halkalı LED'lerden oluşan radyo frekansı (RF) iletişimine dayanan bir sistem ile sağlanmıştır. Sistemin maksimum mesafe hatası 20cm'de %6.5 olarak elde edilirken, maksimum açı hatası 9.4º sapma ile %2.6 olarak elde edilmiştir.

Mesafe ve açı tespit sistemlerinin bir diğer örneği ise küçük robotların birbirlerine göre olan mesafesini ve açısını ölçebilmek için geliştirilen mobil Sesötesi Bağıl Konumlandırma Sistemi (URPS)'dir [9]. Sistem tasarımının arkasındaki mantık, bir vericiden diğer 3 alıcıya sinyalin aktarım sürelerinin değerlendirilmesine dayanmaktadır. Bu çok yönlü sistem, robotların birbirlerine göre olan konumlarını 6.7m mesafede 8mm ve 3º doğruluğu ile tespit edebilmektedir.

Robert ve diğerleri [14] kapalı alanda uçabilen robotların birbirleri arasındaki mesafe, açı ve yükseklik değerlerini hesaplayabilen, kızılötesi 3 boyutlu bağıl konumlandırma sensörü tasarlamıştır.

Geçmiş çalışmalar göstermektedir ki, mesafe, açı ve yönelim değerlerinin ölçümü sürü robot uygulamalarında, robot sistemlerinin koordinasyon halinde etkin bir şekilde çalışabilmeleri için her zaman önemli bir aşama olmuştur. Mesafe ve açı değerlerindeki yüksek hassasiyet, sesötesi sensörler aracılığıyla sağlanabilir; ancak, sesin yayılımına bağlı olarak bu sensörler çoğunlukla yavaş veri yenileme hızına sahiptir. Bağıl konumlandırmada kızılötesi sensör kullanımının avantajı bu sensörlerin yüksek veri güncelleme hızına ve düşük fiyatlı donanıma sahip olmalarından kaynaklanır; fakat bu sensörlerin sisteme uygulanması pahalı ve zahmetli bir iştir. Literatürdeki sürü robotlarında ölçümlerin çoğu kızılötesi, lazer mesafe bulucu ve sesötesi sensörleri ile sağlanmıştır. Genel olarak, bu sensörlerin gerçek zamanlı uygulanması pahalı, zor ve zaman alıcı bir işlemdir. Bu sensörler gürültüye kolayca maruz kalabilir, bu durum da elde edilen verilerin güvenilirliğini azaltabilir.

Bu çalışmada, sürü robotiğinde kullanılabilecek gerçek-zamanlı ve basit, tek kamera ile çalışan bir görüntü algılama sistemi tasarlanmış ve bu sistemin tasarımında robotların birbirlerine göre olan mesafe, açı ve yönelim ölçümlerini tek bir işaretleyici ile alabilmemizi sağlayan çok amaçlı pasif işaretleyiciler temel alınmıştır. Bu sistem, robotlar arasındaki mesafeyi, açıyı ve komşu robotların yönelimini ölçebilen sürü robot sistemleri için tasarlanmış, literatürdeki ilk otonom, basit, güvenilir ve kolay kullanılabilen sistem olarak değerlendirilebilir.

## 2. Yöntem

### 2.1. İşaretleyici Tasarımı

Bu çalışmadaki temel hedeflerden bir tanesi de çok amaçlı bir işaretleyici tasarlamaktır. Bir diğer ifadeyle, tek bir işaretleyici diğer robotların birbirlerine göre olan mesafesi, açısı ve yönelimi hakkında doğru ve güvenilir bilgi sağlamalıdır. Bu işaretleyici, sürü içerisinde birbirlerine göre konumları ve mesafeleri sürekli değişen robotlar için de kullanılabilir olmalıdır. Bu amaçlar doğrultusunda, silindirik yüzey üzerine uygulanan farklı renk ve şekil kombinasyonlar işaretleyici tasarımında baz alınan temel nokta olmuştur. Bunun sebebi, silindirik yüzeyler üzerindeki renkli şekillerin farklı mesafelerden ve doğrultulardan bozulmadan ve dağılmadan okunabilir olmasıdır. İşaretleyiciler, farklı çaplarda (3cm, 4cm ve 5cm gibi) ve yüksekliklerde (5cm, 6cm, 7cm gibi) tasarlanmıştır. Bu çalışmada, işaretleyici desenleri simetrik ve simetrik olmayan desenler olarak iki kategoriye ayrılmıştır. Simetrik desenler ile robotların yönelim açısı saat yönünde ve saat yönünün tersinde olacak şekilde 0 ile 180 derece arasında algılanmıştır. Simetrik olmayan desenler ile ise robotların yönelim açısı 0 ile 360 derece arasında algılanabilmiştir. Şekil 2 ve 3 çalışma esnasında yapılan deneylerde kullanılan bazı işaretleyicileri göstermektedir.

Çalışmada, silindirik işaretleyicilerin üzerinde, iç desen ve dış desen olarak adlandırılan iki farklı desen tasarlanmıştır. Dış desen, silindirin en üst ve alt kısımlarında iki farklı renkten oluşan halka şeklindeki desendir. Dış desenin üst kısmında kırmızı halka ve alt kısmında mavi halka bulunmaktadır. İç desen ise farklı şekil ve renklerin kullanıldığı, yorumlanabilir şekillerdir. Şekil 4 bu çalışmada kullanılan işaretleyiciyi göstermektedir.

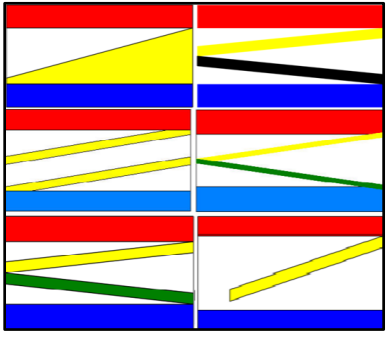

*Şekil 2*: Deneyde uygulanmış olan asimetrik işaretleyiciler.

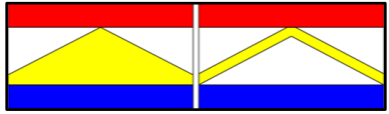

*Şekil 3*: Deneyde uygulanmış olan simetrik işaretleyiciler.

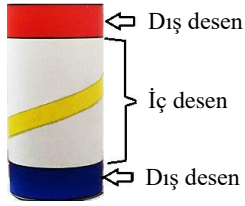

*Şekil 4*: Test edilmiş asimetrik bir işaretleyici modeli.

### 2.2. Görüntü Algılama Algoritmaları

Bu bölümde, mesafe, açı ve yönelim tespiti için önerilen görüntü algılama algoritmaları Python'daki OpenCV kütüphanesi ile tasarlanmış ve uygulamaya geçirilmiştir. Bütün görüntüler gürültüye tabii olacağı için, herhangi bir görüntü algılama algoritması uygulanmadan önce, istenmeyen gürültüler bazı gürültü bastırıcı filtreler ile en aza indirilmiştir. Bu projede, gaussian, medyan ve bilateral olmak üzere farklı boyutlarda üç popüler filtre uygulanmıştır [18]. Bunlar arasından en iyi sonuç, filtre boyutu 15 olan medyan filtre ile elde edilmiştir. Görüntü algılama algoritmasının, renk modeli HSV seçilmiştir, bunun nedeni, ışık değişimine karşı en duyarsız olan renk modelinin HSV olmasıdır [19].

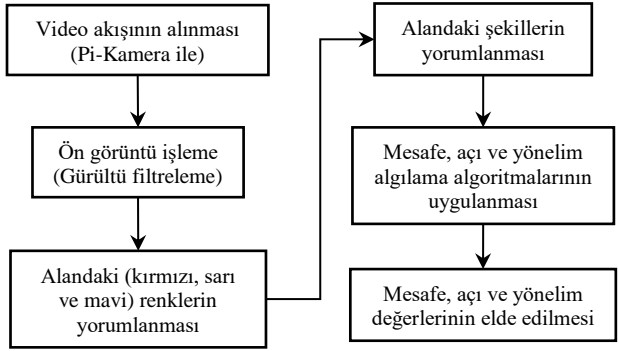

*Şekil 5*: Görüntü algılama sisteminin akış şeması

Önerilen algoritmaların altında yatan mantık; sürü içerisindeki robotların birbirlerine göre olan mesafesini, açısını ve yönelimini hesaplamak için, renk ve geometrik karakteristiklerin belirlenmesi ve yorumlanmasıdır. Tasarlanan görüntü algılama sisteminin akış şeması Şekil 5'te gösterilmiştir.

*2.2.1. Mesafe ve Açı Tespit Algoritması*

Mesafe tespitinde, silindirin alt ve üst kısmında bulunan, kırmızı ve mavi halkalardan oluşan dış desenler kullanılmıştır. Mesafenin ölçümü için önerilen mantık, kırmızı ve mavi bölgelerin merkezleri arasındaki uzaklığı temel almaktadır. Bahsedilen ölçümü yapabilmek için, kırmızı ve mavi alanların yorumlanması ve alan merkezlerinin bulunması gerekmektedir. KOVAN laboratuvarında kontrollü ışık yoğunluğu altında bulunan test alanında, kırmızı ve mavi alanlara ait piksellerin histogramları ve OpenCV'nin interaktif modülü kullanılarak kırmızı ve mavi alanların piksel değerleri yorumlanmış ve iki renkli filtre tasarlanmıştır. Alanların merkez noktaları moment yöntemi kullanılarak hesaplanmıştır [20]. Kullanılan moment denklemi şöyle tanımlanabilir:

$$\mu_{pq} = \sum_x \sum_y (x - \bar{x})^p (y - \bar{y})^q f(x, y) dx dy \quad (1)$$

Moment denkleminden elde edilen, merkez noktaları ise şöyle hesaplanabilir:

$$\bar{x} = \frac{m_{10}}{m_{00}} \quad \text{ve} \quad \bar{y} = \frac{m_{01}}{m_{00}} \quad (2)$$

Merkez koordinatlarının bulunmasından sonra, merkezler aralarındaki mesafe piksel cinsinden hesaplanmıştır. Ardından, robotun kamera lensi ve hedef işaretleyicinin yüzeyi arasındaki mesafe elle ölçülmüştür. Deney, robot ve hedef arasında pek çok mesafeden ve görüş açısından tekrarlanmış ve bütün veriler piksel ve santimetre cinsinden kayıt edilmiştir. Şekil 6 bu deneylerden birini şematik olarak göstermektedir. Şekil 6(a) ve 6(b) robotların birbirlerine göre mesafeleri arasında $L_1>L_2$ ilişkisinin olduğu, robotun iki farklı noktadaki durumunu göstermektedir. Şekil 6(c) ve 6(d) ise sırasıyla Şekil 6(a) ve 6(b)'deki $L_1$ ve $L_2$ değerlerine karşılık gelen piksel cinsinden $P_1$ ve $P_2$ bağıl mesafelerini ve $P_1<P_2$ ilişkisini göstermektedir. Sonuç olarak robot hedefe yaklaştıkça, piksel cinsinden mesafe değerinin arttığı söylenebilir.

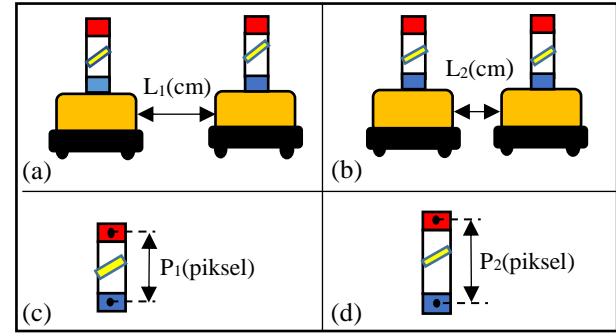

*Şekil 6*: (a) Uzak bir noktadan mesafe hesabı, (b) Yakın bir noktadan mesafe hesabı, (c) a'daki senaryonun piksel değerinden mesafesi, (d) b'deki senaryonun piksel değerinden mesafesi.

Robotun hedefe olan dikey mesafesinin($d_v$) ölçümünden sonra, robotun hedefe göre olan açısını bulmak için bazı geometrik hesaplamalar yapılmıştır. İlk olarak, robotun referans ekseni ile hedef arasındaki yatay dik uzaklık($d_h$) hesaplanmıştır, daha sonra robot ile hedef arasındaki dikey dik

uzaklık($d_v$) kullanılarak geometri formüllerinden robotların birbirlerine göre olan açısı ve mesafesi hesaplanmıştır. Şekil 7 bağıl açı($\theta$), bağıl mesafe($d$) ve yönelim($\psi$) hesabını şematik olarak göstermektedir.

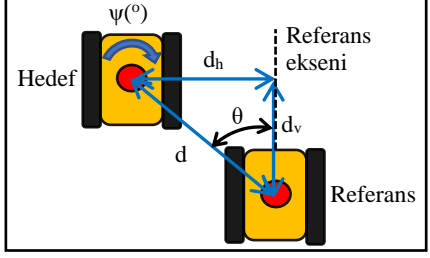

*Şekil 7*: Robotların birbirlerine göre olan açısının tespitinin üstten şematik görünüşü

Şekil 7'ye göre robotların birbirlerine göre olan açısı($\theta$) şöyle hesaplanır:

$$\theta = \operatorname{atan}\left(\frac{d_h}{d_v}\right) \quad (3)$$

Şekil 7'de gösterilen, robotların birbirlerine göre olan mesafesi($d$), hedef ile robotun merkezleri arasındaki yatay dik uzaklık($d_v$) ve bağıl açı($\theta$) kullanılarak şu şekilde hesaplanır:

$$d = \frac{d_v}{\cos\theta} \quad (4)$$

### 2.2.2. Yönelim Tespit Algoritması

Bu algoritmada, renkli iç desenin ve mavi bölgenin merkezi arasındaki uzaklık, robotun sürü içindeki yönelimini elde etmede bir kriter olarak kullanılmıştır. Bu amaçla, Şekil 3 ve 4'te görülen simetrik ve simetrik olmayan 8 farklı desen sisteme uygulanmış ve performansları değerlendirilmiştir. Simetrik olmayan desenler ile 0'dan 360 dereceye kadar olan yönelim değerleri ölçülebilirken, simetrik desenler ile sadece 0'dan 180 dereceye kadar olan değerler ölçülebilmiştir. Araştırmanın amacı 0'dan 360 dereceye kadar olan bütün değerlerini hesaplayabilmek olduğu için, birkaç deneme sonrasında Şekil 2'deki simetrik olmayan modeller arasından sarmal desen, çalışmanın en iyi adayı olarak seçilmiştir.

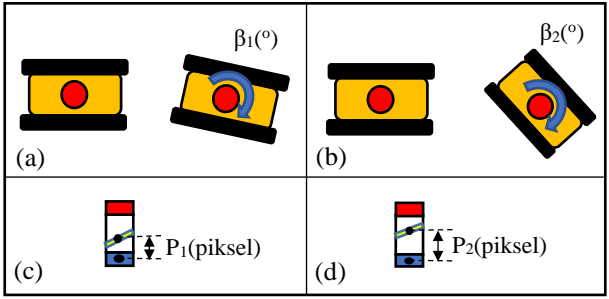

*Şekil 8*: (a) Küçük bir açıda yönelim hesabı, (b) Büyük bir açıda yönelim hesabı, (c) a'daki senaryonun piksel cinsinden yönelim değeri, (d) b'deki senaryonun piksel cinsinden yönelim değeri.

Şekil 8 hedef yöneliminin tespiti için örnek bir senaryoyu göstermektedir. Sarmal desen kullanımındaki mantık, işaretleyici döndükçe sarmal desenin merkezi ile mavi bölgenin merkezi arasındaki değişen uzaklığın, yönelim açısının tespitinde bir kriter olmasıdır. Şekil 8(a) ve 8(b) hedef robotun yönelim açıları arasındaki ilişkinin $\beta_1 > \beta_2$ olduğu iki farklı durumu göstermektedir. Şekil 8(c) ve 8(d) ise sırasıyla Şekil 8(a) ve 8(b)'deki $\beta_1$ ve $\beta_2$ değerlerine karşılık gelen piksel cinsinden $P_1$ ve $P_2$ yönelim değerlerini ve $P_1 < P_2$ ilişkisini göstermektedir. Sonuç olarak yönelim değeri arttıkça, sarmal desen ile mavi bölge arasındaki uzaklığın ya da piksel cinsinden yönelimin arttığı söylenebilir. Şekil 6'daki durumdan farklı olarak, bu durumda yönelimin derece ve piksel değerleri arasında doğrudan bir ilişki olduğu görülmektedir.

## 3. Donanım

Bu kısımda, deneylerde kullanılan donanımsal araçların teknik özellikleri açıklanacaktır. Çalışmalarda robot olarak Pololu Robotik ve Elektronik tarafından geliştirilen Pololu Zumo robot kullanılmıştır [16]. Zumo robotun küçük olması, sürü robotiğinde kullanılan robotları temsil etme açısından uygun olmuştur. Ayrıca, bu robotun temel özelliklerinden bir tanesi de Arduino® ve Raspberry Pi® gibi mikro kontrolcülerle kullanıma uygun olmasıdır. Gerçek-zamanlı görüntü işleme için, 3. nesil küçük bir bilgisayar olarak tanımlanabilecek Raspberry Pi'nin B+ modeli kullanılmıştır [17]. Bu kontrolcüde, 1 GB RAM, 802.11n kablosuz bağlantı modülü ve işlemci olarak 1.2 GHz 64-bit dört çekirdekli ARMv8 kullanılmaktadır. Bunun yanı sıra, 32 GB'lık bir hafıza kartı da sisteme dahil edilmiştir. Bu hafıza görüntünün işlenmesini sağlayan OpenCV modülünün ve Pi-Kamera'nın gerekli yazılımlarının yüklenmesi ve kamera arayüzünün kurulumu için kullanılmıştır.

Pi-Kamera modülünün Rassperry Pi'ye bağlanmasından sonra, gerekli diğer aksamların da Zumo robota yerleştirilmesiyle robot son halini almıştır. Robotun son hali Şekil 9(b)'de sunulmuştur. Pi-Kamera yerden 85mm yüksekliğe yerleştirilmiştir ve deneyler süresince bu mesafe sabit kalmıştır. Kameranın çözünürlüğü 640x460 piksel olarak ayarlanmıştır ve saniyede 32 kare alınabilmiştir. Deneylerin uygulanması için, beyaz arka planlı, 114x63x26 cm boyutlara sahip, özel bir alan tasarlanmıştır. Şekil 9 robotu ve deneylerin yapıldığı alanı göstermektedir.

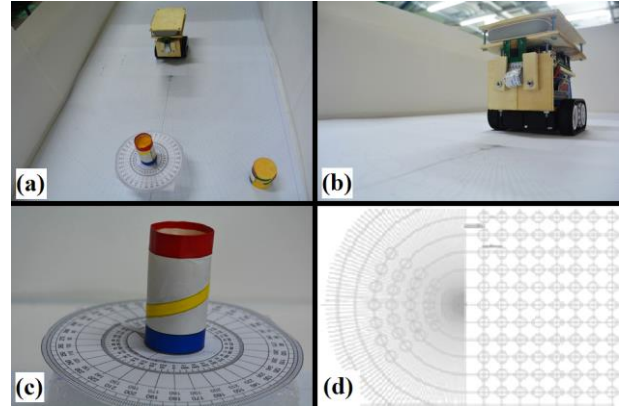

*Şekil 9*: (a) Test düzenek alanı, (b) Robot konfigürasyonu, (c) Örnek bir işaretleyici, (d) Dereceli test düzeneği zemini.

## 4. Tartışma ve Bulgular

Bu çalışmada, test edilmiş 8 farklı işaretleyici modeli arasından Şekil 4'te görülen simetrik olmayan model ele alınmış ve robotların birbirlerine göre olan mesafe, açı ve yönelim değerleri görüntü algılama algoritmaları ile ölçülmüştür. Robota göre ölçülen dikey mesafe, $d_v$ (Şekil 7) verilerinin analizinden sonra, piksel cinsinden dikey mesafe ile cm cinsinden dikey mesafe arasındaki ilişki şu şekilde bulunmuştur:

$$d_v(cm) \propto \frac{k}{d_v(Piksel)} \quad (5)$$

Şekil 10 kaydedilmiş veriden hesaplanmış böyle bir ilişkiyi göstermektedir.

Kaydedilmiş verilerin istatiksel analizinden sonra, dikey mesafenin($d_v$) 9.9cm'den 70cm'ye değiştiği aralıkta, k sabiti 3193.7 olarak hesaplanmış ve ilgili denklem 5 şu şekilde güncellenmiştir:

$$d_v(cm) = \frac{3193.7}{d_v(Piksel)} \quad (6)$$

Denklem 5 ve 6'ya göre, işaretleyici üzerindeki üst kırmızı halka ve alt mavi halka arasındaki ölçülen piksel mesafesi büyüdükçe, robot ve hedef arasındaki cm cinsinden dikey mesafe azalır, dolayısıyla bu yol ile sürü içerisindeki robotlar arasında olabilecek olası çarpışmalar önlenebilir.

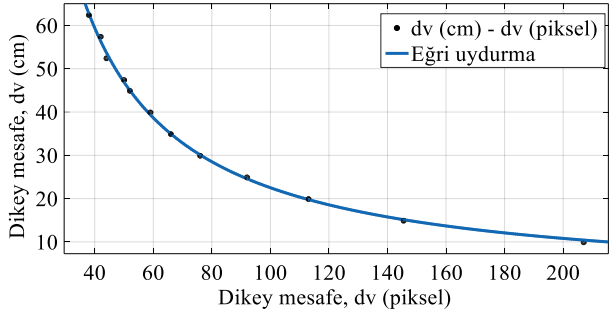

*Şekil 10*: Dikey mesafe($d_v$) hesabında piksel - cm ilişkisi.

Hedef yönelimini ölçmek için tasarlanmış olan görüntü algılama algoritması, sarı renkli sarmal desenin ve mavi bölgenin merkezlerini tespit etmiş ve iki nokta arasındaki mesafeyi hesaplamıştır. Sarmal desenlerin geometrisinden dolayı, işaretleyici döndükçe, tespit edilmiş mesafe de değer olarak piksel biriminde değişmiştir. Robot ve hedef arasındaki dikey uzaklık($d_v$) sadece 12cm ve 37cm aralığında değiştiğinde, sarı ve mavi bölgeler arasındaki mesafe ölçülebilmiştir. Robot 12cm'den 37cm'ye hareket ettikçe, bu aralıktaki her 1 cm için, sarmalın merkezi ile mavi bölgenin merkezi arasındaki piksel cinsinden dikey mesafe ilişkisinin doğrusal olduğunu anlaşılmıştır. Bu ilişki kullanılarak hedef robotun yönelim değeri($\psi$) şu şekilde elde edilmiştir:

$$\psi = k_1 * (d_v\_piksel) - k_2 \quad (7)$$

İstatiksel verinin belirtilen mesafe aralığında manipüle edilmesi, $k_1$ ve $k_2$ sabitlerinin Şekil 11'deki gibi değiştiğini göstermiştir. Şekil 11'e göre, 12cm'den 37cm'ye kadar olan her mesafe için, $k_1$ doğrusal olarak 2.0951'den 6.2095'e değişirken, $k_2$ de üstel olarak 97.179'dan 94.421'e düşmektedir.

Robotların birbirlerine göre olan açısı, robotlar arasındaki dikey($d_v$) ve yatay($d_h$) mesafelerin (Şekil 7) ölçümünün ardından, 3 numaralı denklem kullanılarak hesaplanabilmiştir. Deneylere göre, bağıl açılar ancak robotun hedeften yatay mesafesi 5.1cm ile 23cm aralığında yer aldığında saptanabilmiştir. Kaydedilen verilerin istatistiksel analizi $d_h$(piksel) ve $d_h$(cm) arasındaki ilişkiyi göstermiştir. Bu doğrultuda, her $d_h$ mesafesinde, cm ve piksel cinsinden değerler arasındaki ilişki aşağıdaki şekilde tanımlanabilir:

$$d_h(cm) = k * d_h(Piksel) \quad (8)$$

Bu denklemde, k sabiti, robot ve hedef arasındaki yatay mesafenin 5.1cm ile 23cm arasında değişmesine göre 0.0206 ile 0.0764 arasında değişmektedir. Bu ilişki Şekil 12'de gösterilmektedir.

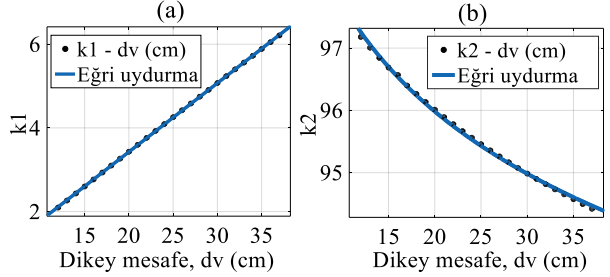

*Şekil 11*: (a) 12cm ile 37cm arası değişen dikey mesafe ve $k_1$ arasındaki ilişki, (b) Dikey mesafe ve $k_2$ arasındaki ilişki

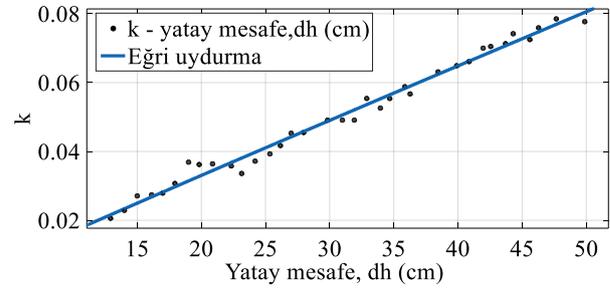

*Şekil 12:* Yatay mesafe($d_h$) ve k terimleri arasındaki ilişki.

Denklem 6'dan $d_v$ mesafesinin ve denklem 8'den de $d_h$ mesafesinin hesaplanmasından sonra, denklem 3 kullanılarak robotların birbirlerine göre olan açısı doğrudan hesaplanabilmiştir. Önerilen görüntü algılama algoritmalarının gerçekçi koşullarda test edilmesi sonucunda, sistemin yönelim değerlerinin belirlenmesindeki hatası %3.25 ve standart sapması 0.0289 olarak hesaplanmıştır. Bağıl açıların hesaplanmasında ise sistemin yüzdelik hatası 0.0597 standart sapma ile %4.044 olarak bulunmuştur. Sonuç olarak, gerçekçi koşullardaki testler ve istatistiksel analizler, sistemin bağıl açı ve yönelim değerlerinin belirlenmesindeki yüksek performansını kanıtlamıştır.

Robotların birbirlerine göre olan mesafesi, bağıl açının ölçümünün ardından, 4 numaralı denklem kullanılarak hesaplanabilmiştir. Değişik ışık koşullarında ve farklı mesafelerde yapılan deneyler sonucunda, bağıl mesafe tespitinde sistemin hata yüzdesi 0.0044 standart sapma ile %0.414 olarak elde edilmiştir. Elde edilen veriler, uygulanan sistemin güvenilir ve etkin olduğunu göstermektedir.

*Tablo 1*: Değişkenlerin tespitindeki hata ve sınırlamalar

| Tespit Edilen Değişkenler | Kabul Edilebilir Sınırlar | | En Yüksek Hata Oranı |
|---|---|---|---|
| | Dikey Mesafe ($d_v$) | Yatay Mesafe ($d_h$) | |
| Bağıl mesafe (d) | 13-50 cm | - | %0.414 |
| Bağıl açı (θ) | 13-50 cm | 5.1-23 cm | %4.044 |
| Yönelim (ψ) | 12-37 cm | - | %3.25 |

Diğer fiziksel sistemlere benzer olarak, önerilen algoritmalar, bazı sınır değerleri ile kısıtlanmaktadır. Sistem, bağıl mesafeyi ve açıyı, robot ile hedef arasındaki dikey mesafe sadece 13cm ile 50cm arasında değiştiğinde belirleyebilmektedir. Buna ek olarak, yönelim değerleri için 0° ile 45° arasında aykırılık bölgesi bulunmaktadır. Diğer bir deyişle, hedef belirtilen aralıkta yer aldığında, robot yönelim açısını belirleyememektedir.

Tablo 1 elde edilen bulgulardaki hata ve sınırlamaları göstermektedir.

## 5. Sonuç

Bu çalışmada, sürü içerisindeki küçük robotların bağıl mesafe, açı ve yönelim değerlerinin tespit edilmesini sağlayan görüntü algılama algoritmaları önerilmiş ve Raspberry Pi tabanlı gerçek zamanlı görüntü işleyebilen mobil bir sistem üzerinde uygulanmıştır. Lazer mesafe bulucu, sesötesi ve kızılötesi gibi aktif sensörler, stereo görüş gibi derinlemesine görüntü algılama sensörleri ve diğer pahalı ve karmaşık metotlarla karşılaştırılınca, bu çalışmada önerilen algoritmanın mesafe, açı ve yönelim değerlerini, Pi-Kamera'nın 2 boyutlu dijital görüntülerini kullanarak ölçebilen sürü robot sistemleri için tasarlanmış özgün bir metot olduğu söylenebilir. Buna istinaden, 3 boyutlu algılama sistemlerinin aksine, bu çalışmada sunulan sistem daha az hafıza kullanmakta ve gerçek zamanlı işleme sistemlerine daha etkin uygulanabilmektedir. Elektrik enerjisi kullanan, pahalı ve test ortamında kurulması zor olan LED gibi aktif işaretleyiciler ile karşılaştırıldığında ise bu çalışmada her türlü baskı yapılabilen malzemelerden oluşan daha basit pasif işaretleyiciler kullanıldığından dolayı, uygulanan görüntü algılama sistemi ayrıca maliyet açısından daha uygundur. Sistemin bahsedilen 3 farklı parametresinin; bağıl mesafenin, açının ve yönelimin aynı anda tek bir işaretleyici kullanılarak ölçülebilmesi, tasarlanan işaretleyiciyi bu çalışmanın öne çıkan bir özelliği yapmaktadır. Gelecekte, bu çalışma çoklu robot senaryoları için geliştirilecek ve test edilecektir. Ayrıca, sistemin performansının, doğruluğunun ve hızının geliştirilmesi de dikkate alınacaktır. Çalışmada gerçekleştirilen gerçek zamanlı görüntü işlemeyi ve robotun analiz ettiği sahneyi içeren bir testin örnek videosu [21]'de bulunabilir.

## 6. Kaynakça